\documentclass[final]{cvpr}

\usepackage{times}
\usepackage{epsfig}
\usepackage{graphicx}
\usepackage{amsmath}
\usepackage{amssymb}
\usepackage{dsfont}

\usepackage[pagebackref=true,breaklinks=true,colorlinks,bookmarks=false]{hyperref}

\def\cvprPaperID{29} 
\def\confYear{CVPR 2021}

\begin{document}

\title{Shadow Neural Radiance Fields for Multi-view Satellite Photogrammetry}

\author{Dawa Derksen\\
European Space Agency - ESTEC\\
Keplerlaan 1, 2201 AZ Noordwijk, Netherlands\\
{\tt\small dawa.derksen@esa.int}
\and
Dario Izzo\\
}

\maketitle

\begin{abstract}
We present a new generic method for shadow-aware multi-view satellite photogrammetry of Earth Observation scenes. Our proposed method, the Shadow Neural Radiance Field (S-NeRF) follows recent advances in implicit volumetric representation learning. For each scene, we train S-NeRF using very high spatial resolution optical images taken from known viewing angles. The learning requires no labels or shape priors: it is self-supervised by an image reconstruction loss. To accommodate for changing light source conditions both from a directional light source (the Sun) and a diffuse light source (the sky), we extend the NeRF approach in two ways. First, direct illumination from the Sun is modeled via a local light source visibility field. Second, indirect illumination from a diffuse light source is learned as a non-local color field as a function of the position of the Sun. Quantitatively, the combination of these factors reduces the altitude and color errors in shaded areas, compared to NeRF. The S-NeRF methodology not only performs novel view synthesis and full 3D shape estimation, it also enables shadow detection, albedo synthesis, and transient object filtering, without any explicit shape supervision.
\end{abstract}

\section{Introduction}
Many people think of satellite images as being from straight above, but they are almost always taken at an oblique angle. So-called \emph{off-nadir} images provide information regarding vertical structure, and can be useful for estimating 3D shape. Moreover, the combination of images from different viewpoints reveals aspects that are in most cases impossible to capture with only one image.

One of the goals of multi-view imagery missions (SPOT6-7, WorldView-3) is to estimate the topography of the Earth's land cover. This knowledge can be relevant for several applications in Remote Sensing, including terrain mapping for flood risk mitigation~\cite{mcclean2020implications}, biomass estimation~\cite{simard2006mapping}, land cover classification~\cite{demarez2019season} and change detection~\cite{qin20163d}.

While it is possible to use an active sensor such as a Light Detection And Ranging (LiDAR) to directly measure the distance from the satellite to the surface, these require significant amounts of energy compared to passive cameras.

\begin{figure}[h!]
\begin{center}
   \includegraphics[width=.9\linewidth, page=4, trim={0 0 12cm 0}]{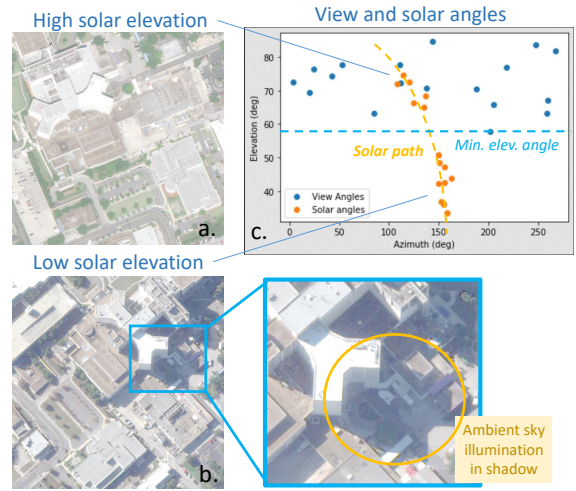}
\end{center}
\caption{\label{fig:angles}Two off-nadir WorldView-3 images with high (a.) and low (b.) solar elevation ($\theta_s$). Strong temporal non-correlation principally manifests as changing shadows. The zoom on image b. shows that shadows are weakly illuminated by diffuse blue light coming from the sky. The graph (c.) displays the distribution of viewing angles (blue) and solar angles (orange). The solar directions are concentrated around the solar path, illustrated by the dotted orange line. They are clustered in two groups with no images in between.}
\end{figure}

In this research we focus on the performances achievable by optical sensors alone, motivated by applications with strong restrictions on available energy. These could include micro/nano-satellites for Earth Observation, or probes for space exploration around other planets~\cite{gwinner2016high}, comets or asteroids. 

\begin{figure*}
\begin{center}
   \includegraphics[width=.9\linewidth, page=1, trim={0 4cm 6cm 0.5cm}]{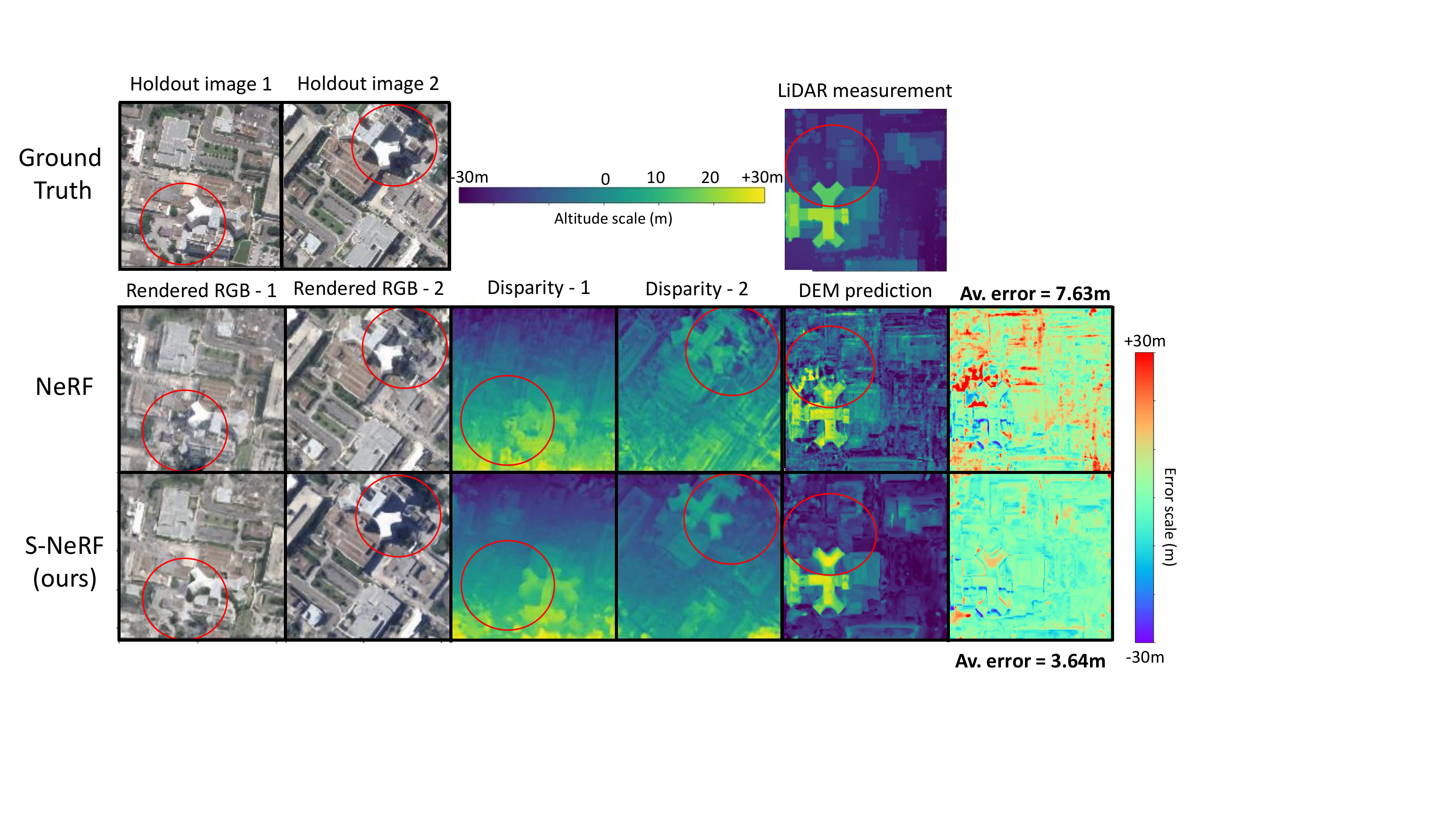}
\end{center}
\caption{\label{fig:nerf}Comparison between NeRF~\cite{mildenhall2020nerf} and S-NeRF. The first line displays two sources of ground truth information; test images in unseen viewing and lighting conditions, and a LiDAR measurement of the surface altitude. The second line demonstrates the performance of NeRF on both novel view synthesis and surface altitude prediction (DEM). The circled area, susceptible to containing non-correlation due to shadows, is where most of the errors are found both in shape and color. Our results are shown in the third row. S-NeRF allows for a higher quality estimation of the shadows, and produces a more accurate altitude prediction than NeRF. This is evident in the last column which shows the difference between the measured and estimated altitudes as well as the average absolute altitude error.}
\end{figure*}

In this paper, we rework one of the latest methods in 3D computer graphics, Neural Radiance Fields~\cite{mildenhall2020nerf} (NeRF) to perform photogrammetry of urban scenes using Very High Spatial Resolution (VHSR) optical images from off-nadir viewing angles. Our problem differs from on-ground computer vision applications in several ways. 

First of all, the images exhibit strong \emph{non-correlation} because they are taken at different moments in time. Figure~\ref{fig:angles} shows an example of non-correlation due to varying lighting conditions on two WorldView-3 images~\cite{le2019ieee}. Shadows moving in between images can cause errors to manifest when using altitude estimation methods based on pixel or feature matching~\cite{rupnik20183d, rupnik2019more,krauss2019cross}.  Moreover, shadows cannot be easily detected due to ambient illumination from the sky. A further discussion about the different non-correlation sources and their impact on photogrammetry is made in Section~\ref{sec:sgm}.

Second, the number and distribution of viewing angles is restricted. A yearly WorldView-3 data set such as the one used in our experiments contains in average 10-20 images with an off-nadir angle lower than $35^{\circ}$. Figure~\ref{fig:angles} also reveals how the distribution of solar angles is relatively sparse, because the data was originally captured with stereo pair/triplet processing in mind.

The latest research in image-based photogrammetry has seen the arrival of a new family of methods, known as Neural Radiance Fields (NeRF)~\cite{mildenhall2020nerf}. These have recently been explored for computer vision tasks such as novel view synthesis, which is the process of creating images of an object from an unseen viewing angle, given a set of 2D images of the object. By learning an internal representation of the volume as a continuous 3D function, NeRFs succeed in generating photo-realistic images from previously unseen viewing directions, using 20-100 posed training images. More details are provided in Section \ref{sec:nerf}.

We demonstrate that the NeRF approach shows errors when applied to images with inconsistent lighting sources in Figure~\ref{fig:nerf}. An explicit model of the shadow effects is necessary in order to properly learn the shape of the scene, when faced with non-correlation due to changing lighting conditions. 

This paper presents Shadow Neural Radiance Fields (S-NeRF), an extension of NeRF~\cite{mildenhall2020nerf}, that enables their application to multi-view satellite imagery with varying light conditions. Our changes to the NeRF model and rendering scheme are detailed in Section \ref{sec:method}.

\begin{itemize}
    \item[] \textbf{Section} \ref{sec:incoming} - Incoming light from a directional white light source (the Sun) and a diffuse colored light source (the sky) are learned by the model for each scene, in order to render realistic shadows with ambient light effects.
    \item[] \textbf{Section} \ref{sec:light} An additional training step that traces rays from the light source into the scene is proposed to improve the quality of the learned light source visibility function at each point.
    \item[] \textbf{Section} \ref{sec:altitude} An altitude-based sampling scheme is employed to better fit the overall flat shape of Earth Observation scenes.
\end{itemize}

In Section \ref{sec:experiments}, we demonstrate the performance of S-NeRF on two tasks : novel view synthesis in unseen light conditions, and altitude estimation. We quantitatively show that S-NeRF outperforms NeRF on both tasks. We also show how using an explicit shadow model allows us to detect shadows in the training and test images, and estimate the albedo on the extracted 3D surface, even in areas that are persistently in the shadow in the images.

\section{Related work}

Learning the three dimensional properties of a scene based on two dimensional image information alone is an important challenge for Computer Vision and Remote Sensing alike. This task is often referred to as inverse rendering. Generally speaking, the objective is to estimate physical aspects such as opacity, spectral albedo, roughness, metallicity, or others, in the 3D volume or on a 2D subspace at the surface of the object. In theory, inverse rendering appears as a fundamentally under-constrained problem, an illustration is the monochrome scene, where the absence of visual cues means an infinity of possible shapes could be equally valid explanations for any set of images. In most natural images, however, the presence of texture and sharp details are sufficient to infer three dimensional structure with a reasonable number of observations.

Automatic inverse rending methods that reason about the 3D nature of the scene can be divided into the following groups, which differ on the kind of data that they require to operate.

Methods of the first group require prior knowledge of the shape of the observed scene. With this, it becomes relatively straightforward to project the different images onto the shape, to learn aspects such as surface specularity and shadows, to perform high-quality relighting~\cite{martin2020gelato, oechsle2020learning}, or to refine the 3D geometry~\cite{bittner2018automatic}.

The second group seek to match visual patterns in the images with 3D shape, by training on examples of images where the scene shape is known. Once training has been accomplished, the models are able to infer the 3D shape of a new scene using a stereo pair~\cite{zbontar2016stereo}, or a single image~\cite{mou2018im2height,carvalho2020multitask}. However, this requires large data sets of images with a known shape for training.

The third group, which are the focus of this paper, are ``unsupervised'' or rather \emph{self-supervised}: they infer the shape and surface properties from the images only. For Earth Observation with optical sensors alone, and for many other space-based applications, this appears to be the most interesting approach as it does not require any external form of data regarding surface shape. These methods are particularly relevant for space exploration where very limited prior shape information is available. 

Section~\ref{sec:sgm} elaborates upon methods that explicitly match different image elements together~\cite{schonberger2016structure, hirschmuller2007stereo}. Section~\ref{sec:nerf} shows recent progress made in computer vision based on light transport through an implicit volumetric representation~\cite{lombardi2019neural, mildenhall2020nerf, bi2020neural}. 

\subsection{Stereo Matching}
\label{sec:sgm}
Many efforts in the direction of satellite-based photogrammetry have been focused on Stereo Matching~\cite{de2014automatic,knobelreiter2017end,rupnik2019more,perko2019mapping}. In this paradigm, the surface is represented by a function of the type $f(x,y) = h$, called the Digital Elevation Map (DEM), where $(x,y)$ are spatial coordinates on the Earth (latitude, longitude for example) and $h$ the surface height. These methods explicitly match pairs or triplets of pixels, with the idea that a group of matching rays must intersect at the surface of an object in the scene. This is posed as an optimization problem which maximizes multi-view consistency through mutual information pixel matching, while minimizing a global energy function to encourage regular 3D structures. 

In its basic form, Stereo Matching lacks a way of modeling inconsistencies in the captured scenes, due to the fact that the visual cost is based on feature matching. For example, the movement of shadows between images at different moments of the day can make it difficult to accurately match spatial features. Other sources of temporal non-correlation include specular effects, transient objects (e.g. cars), vegetation growth, land cover change, and weather phenomena (e.g. snow). The impact of these inconsistencies is reduced by using pairs or triplets of images in similar conditions: roughly at the same time of day, and of the year. Studies by~\cite{rupnik20183d, rupnik2019more} adapt the cost function of the semi-global optimization to tolerate non-correlation, but such approaches would most likely fail on strongly non-correlated images, for example, a set of morning and afternoon images mixed together. 

\subsection{Neural Radiance Fields}
\label{sec:nerf}
Neural Radiance Fields (NeRF)~\cite{mildenhall2020nerf}, simultaneously model color and geometry using a volumetric representation written as:
$$
f(\mathbf{x}) = (\mathbf{c_{\lambda}}, \sigma)
$$

The representation function $f$ is defined on 3D space by the coordinate vector $\mathbf{x}\in \mathds{R}^3$ The color of outgoing light $\mathbf{c_\lambda}$, also called \emph{radiance}, is defined at discrete wavelengths: for visible images, $\mathbf{c_\lambda} \in [0, 1]^3$. The shape is modeled by $\sigma \in \mathds{R}^+$, the density, which expresses how transparent or opaque a medium can be. In this study, we focus only on the effects of shadows on a Lambertian scene, and therefore omit the dependency of the radiance on the viewing angles. The combination of these factors is discussed in Section \ref{sec:discussion}.

NeRFs construct $f$ using a Multi-Layer Perceptron (MLP) network, thereby achieving a continuous ``implicit representation'' which has more interesting scaling properties than a voxel grid~\cite{lombardi2019neural, bi2020deep}. The differentiable nature of the neural network is also key to solving the inverse rendering problem using gradient descent.

\subsubsection{Fully emissive rendering model}
Optical volume rendering works by simulating light transport through a 3D representation, to infer what image a camera would see from a certain position in space. Each of the pixels in a desired image defines a ray, which accumulates light as it traverses the scene.

Previous successful neural volume rendering approaches~\cite{lombardi2019neural, mildenhall2020nerf} are based on the fully emissive light transport model~\cite{max1995optical}. In practice, the continuous rendering integral is estimated as a discrete sum, by performing a Monte-Carlo integration~\cite{mildenhall2020nerf}. This means sampling the volume properties $(\mathbf{c}, \sigma)$ predicted by the neural network at $N_s$ locations along each ray. The discretized form of the rendering integral is shown in Equation~\ref{eq:alpha_comp}.

\begin{equation}
\begin{split}
    \label{eq:alpha_comp}
    \mathbf{\hat{I}} & = \sum_{i=1}^{N_s}w_i \mathbf{c_i} \\
    w_i & = T_i \alpha_i\\
    \alpha_i & = 1-e^{-\sigma_i \delta x_i} \\
    T_i & = \prod_{j=0}^{i-1}(1-\alpha_j) \\
\end{split}
\end{equation}

The approximated perceived light color $\mathbf{\hat{I}}$ is written as a weighted sum of the color vectors $\mathbf{c_i}$, with weights $w_i$. This numerical integration scheme defines $\alpha_i$ as the opacity along ray segment $i$ of length $\delta x_i$, by definition $\alpha_i \in [0,1]$. The transparency $T_i$ is written as the cumulative product of the inverse opacity, from the origin to the current segment $i$. A high $w_i$ value indicates that ray segment $i$ is either emitting or reflecting light (high $\alpha_i$), and not blocked earlier along the ray (high $T_i$). 

\subsubsection{Training}

Solving the inverse rendering problem means learning how much each individual element of the scene contributes to the final perceived color.

To train a neural volumetric representation to fit with a data set of images, a batch of $N_b$ pixels is randomly sampled from all of the images. The perceived color $\mathbf{\hat I(r)}$ is then estimated along the corresponding rays. This results in a loss value for each pixel, which is back-propagated back to the weights of the neural network. In this way, the neural volumetric representation is modified to better fit with what is observed in the training images. It should be noted that such gradient descent optimization is only possible because both the representation and the rendering functions are differentiable. 

\subsubsection{Neural Reflectance Fields}

For 3D scene relighting, Neural Reflectance Fields~\cite{bi2020neural} were recently introduced. These methods actively reason about the quantity of incoming light at a spatial position, by computing the visibility between each query point and the light source on the fly during inference. Training a Neural Reflectance Field using images with a known non-collocated light source, or multiple light sources, would imply a \emph{quadratic} or even \emph{cubic} increase in the number of sample points required to estimate the light transport function, depending on the complexity of the lighting effects that are modeled. This would be prohibitive for practical purposes, especially in terms of memory. Therefore, Neural Reflectance Fields can only be trained on images taken with a white light source collocated with the viewing axis (camera flash only). Nonetheless, they show success in extracting surface parameters such as the albedo, and perform accurate relighting on objects that contain complex structure such as fine elements. 

Our work proposes a different strategy than Neural Reflectance Fields~\cite{bi2020neural}, which is necessary because in multi-view satellite images (and in many other cases), the light source is non-collocated with the camera axis. For this reason, we propose to restrict the lighting conditions to the two sources that illuminate the Earth: the Sun and the sky. Explicit knowledge of the solar direction is used to simplify the full optical volume rendering formulation. In our method we put high priority on maintaining linear complexity with respect to the number of sample points along the rays, $O(N_s)$ which is \emph{the same complexity as NeRF}.

\section{Method}
\label{sec:method}

\subsection{Irradiance model}
\label{sec:incoming}

We define the spectral irradiance vector $\mathbf{\ell}_{\lambda}(\mathbf{x}, \mathbf{\omega_s})$, as the intensity of incoming light at a 3D location, for the solar direction $\mathbf{\omega_s}$, at wavelengths ${\lambda} \in \{R,G,B\}$.

Instead of explicitly calculating the irradiance at each point \cite{bi2020deep,bi2020neural}, we propose to implicitly model this quantity in the neural network. The main goal of this is to avoid numerically estimating a full 2D or 3D integral during training and inference.

In our method, we model variations in incoming light by introducing two new outputs to the network, $s(\mathbf{x},\mathbf{\omega_s})$ and $\mathbf{sky(\omega_s)}$, which both depend on one new 2D input, the solar direction $\mathbf{\omega_s} = (\theta_s, \phi_s)$. 

Physically, $s(\mathbf{x},\mathbf{\omega_s})$ represents the ratio of incoming solar light with respect to the diffuse sky light. This quantity can also be loosely interpreted as the visibility of the directional light source at 3D location $\mathbf{x}$, along the direction $\mathbf{\omega_s}$. A value of 0 means no solar visibility and 1 means full solar visibility.

The other quantity, $\mathbf{sky(\omega_s)}$ is a learned vector that expresses the color of the illumination incoming from the sky (3D for RGB images). The network predicts these values based on the solar direction alone, and not on spatial coordinates, to model how the sky acts as a diffuse light source that only visibly contributes to lighting up areas in the shadow.

In order to take both of these quantities into account in the rendering, a new term is added to the alpha-compositing formula from Equation~\ref{eq:alpha_comp}.

We define the total irradiance $\mathbf{\ell}$ as a weighted sum of the known and learned light sources, using the network-predicted values of $s$ and $\mathbf{sky}$. The light mixing model, shown in Equation~\ref{eq:light_mix} poses $\mathbf{\ell}$ as a weighted sum of a white light source $\mathds{1}_3$ and the learned sky color using $s$ and $(1-s)$ as weights. This follows from the assumption that the Sun emits white light in the visible bands. 

Instead of directly modeling the radiance $\mathbf{c}(\mathbf{x})$ as was done previously~\cite{lombardi2019neural,mildenhall2020nerf} we explicitly model the albedo $\mathbf{a}(\mathbf{x})$ as a network output (Figure~\ref{fig:archi}). We then use a simple Lambertian reflectance model that dictates that the radiance (noted $\mathbf{c}$ in Equation~\ref{eq:alpha_comp}) is the point-wise product of the irradiance $\mathbf{\ell}$ and the albedo $\mathbf{a}$ vectors.

\begin{equation}
\begin{split}
    \label{eq:light_mix}
    &\mathbf{\ell} = s\mathds{1}_3+(1-s)\mathbf{sky}\\
    &\mathbf{c(\mathbf{x},\mathbf{\omega_s})} = \mathbf{a(\mathbf{x})\cdot{}\mathbf{\ell}(\mathbf{x},\mathbf{\omega_s})}
\end{split}
\end{equation}

\begin{figure}
\begin{center}
   \includegraphics[width=.9\linewidth, page=3, trim={1cm 6.5cm 14cm 0.5cm}]{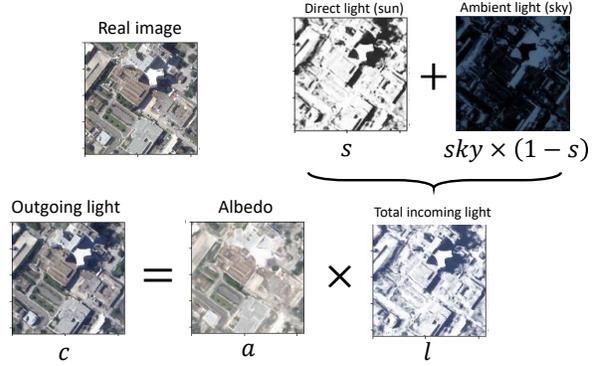}
\end{center}
\caption{\label{fig:render} Schematic of the S-NeRF shading model (Equation~\ref{eq:light_mix}). The outgoing light (radiance) $\mathbf{c}$ at each point along a ray is modeled as a point-wise product between the albedo $\mathbf{a}$ and the total incoming light (irradiance) $\mathbf{\ell}$. The latter is a mixture between the direct white light from the Sun and the indirect light from the sky, which is weighted by the solar visibility function. We see how the faint sky light is learned as a bluish tone, and how the albedo image contains no visible shadows.}
\end{figure}

The new alpha-compositing model to estimate the perceived color \emph{with shading}, $\mathbf{\hat{I}_s}$ is expressed in Equation~\ref{eq:alpha_comp_sh}, using the same definition for $w_i$ as Equation~\ref{eq:alpha_comp}. 

\begin{equation}
\label{eq:alpha_comp_sh}
    \mathbf{\hat{I}_s} = \sum_{i=1}^{N_s}w_i \mathbf{a_i}\mathbf{\ell_i}
\end{equation}
A visual example of the S-NeRF rendering procedure that is described in Equations~\ref{eq:light_mix} and~\ref{eq:alpha_comp_sh} is shown in Figure~\ref{fig:render}. 

This simplification of the full light rendering model avoids extra sampling during training iterations, but does not account for specular effects and makes heavy simplifications on the indirect illumination. In our model, every area in the shadow receives an equal amount of light from the sky, meaning that occlusion of sky light by the scene itself is not taken into consideration. Furthermore, it does not allow for indirect light coming from other parts of the scene, also known as double-bounce indirect illumination. These are very challenging to take into account without increasing the complexity of the integration.

\subsection{Solar correction rays}
\label{sec:light}
To accurately learn $s$, the solar visibility function, we propose to draw a second set of rays at each training iteration. We call these Solar Correction (SC) rays, as they aim to correct the solar visibility function based on the current estimated scene geometry. Along such a ray we compute the transparency $T_i$ from Equation~\ref{eq:alpha_comp}. We then add a term to the optimization loss function to penalize differences between $T_i$ and $s_i$. The idea of this loss is that along a solar ray, the transparency at each point along the ray is equivalent to the ratio of incoming solar light at that point. 

Equation~\ref{eq:loss} shows the loss function $L$ for a batch of rays $b$, which contains three terms.

\begin{equation}
\begin{split}
\label{eq:loss}
    L(b) = & \sum_{\mathbf{r}\in b}||\mathbf{I}(\mathbf{r}) - \mathbf{\hat{I}_s}(\mathbf{r})||_2^2 +\\ & \lambda_s\sum_{\mathbf{r}\in SC}(\sum_{i=1}^{N_s}(T_i-s_i)^2 + 1 - \sum_{i=1}^{N_s}w_is_i)
\end{split}
\end{equation}

\begin{enumerate}
    \item The pixel-based RGB loss, a mean squared error between the network predicted pixels $\mathbf{\hat{I}_s}$ and the image pixels $\mathbf{I}$, of a batch of rays $b$.
    \item The $L_2$ loss between $T$ and $s$, on the solar correction rays (SC), using the same definition for T as Equation \ref{eq:alpha_comp}. To encourage the $s$ values to become close to the current $T$ values, we stop the gradient on $T$.
    \item The $L_1$ norm of $ws$ subtracted from 1. This term expresses the idea that the entirety of the solar light should be absorbed by the visible surface. The gradient of $w$ is stopped. 
\end{enumerate}

$\lambda_s$ is a weight parameter to balance the participation of the different losses. This parameter represents the trade-off between color accuracy and shadow validity. We found $\lambda_s = 0.05$ to work best in our experiments. 


Solar correction rays can be randomly drawn from the upper hemisphere, or focused on the positions of certain requested angles. For the experiments, we used solar correction rays generated from directions along an interpolation axis that follows the solar path (Figure~\ref{fig:angles}). The result of this interpolation, Figure~\ref{fig:interp}, demonstrates the importance of solar correction when generalizing to lighting conditions that are significantly different from the conditions present in the data set.

\subsection{Altitude-based sampling}
\label{sec:altitude}
We define a new sampling scheme for selecting integration points along the rays, to better fit Earth Observation scenes where the bounds of the X and Y dimensions often exceed the bounds of the Z axis. The samples are selected in regular intervals along the absolute Z axis rather than along the sensor axis.

We also sample between minimum and maximum altitude values rather than near and far distances, to avoid introducing parameters which would depend on the satellite configuration. Selecting the minimal and maximal altitude parameters for a given zone can be done in various ways, for example, from a large-scale Digital Terrain Map based on a low-resolution data source. In our case we selected values based on the minimum and maximum values of the airborne LiDAR maps (Table~\ref{tab:results}, 3rd row).

We also use importance sampling as proposed by~\cite{mildenhall2020nerf} on the visual rays only, as it appears to increase the effective depth resolution for a fixed sampling budget.

\subsection{Network architecture}
\label{sec:archi}
\begin{figure}
\begin{center}
   \includegraphics[width=.8\linewidth, page=2, trim={0 3.5cm 16cm 0.5cm}]{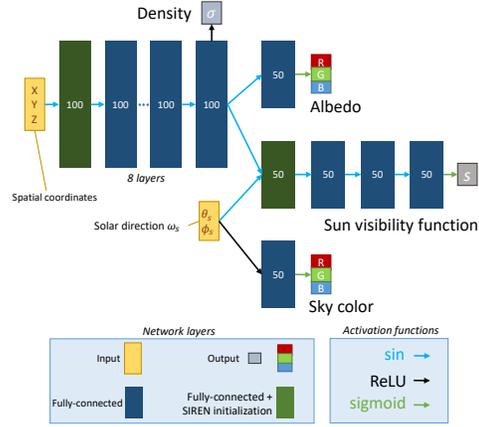}
\end{center}
\caption{\label{fig:archi} S-NeRF network architecture. Spatial coordinates are used as inputs to the density and albedo layers in the same way as the original NeRF~\cite{mildenhall2020nerf}. Our major changes are the addition of 1. a solar visibility network which takes as an input both the (unactivated) outputs of the last density layer and the solar direction $\omega_s$, and 2. a sky color estimation layer.}
\end{figure}

These experiments are based on the network architecture shown in Figure~\ref{fig:archi}. The internal representation network can be divided into four parts.

First, the density output is connected to the spatial inputs through a 8 fully-connected SIREN layers of width 100, using the initialization procedure described by~\cite{sitzmann2020implicit} instead of positional encoding. Importantly, like in~\cite{mildenhall2020nerf}, the final activation function that provides $\sigma$ is a ReLU, so as to provide higher values than 1. We also add noise to the unactivated output with a standard deviation of 10, which we decrease to 0 throughout the training~\cite{schwarz2020graf}.

Second, the RGB albedo is connected to the last layer of the density network, activated by a sigmoid. Next, the solar visibility $s$ is connected to the last layer of the density network, concatenated to the solar direction. Here, 4 layers are used, with initial output noise standard deviation of 1. Finally, the sky color is estimated using only the solar direction, through 1 ReLU layer. 

Explicitly separating the contributions of these four factors enables an independent synthesis of each property, which can help in correcting them individually (solar correction rays), or to render certain properties without others (shadow-free rendering).

\section{Experiments}
\label{sec:experiments}
We evaluate the performance of S-NeRF on satellite imagery from the WorldView-3 optical sensor~\cite{le2019ieee}, training the model using only the visible RGB bands. Instead of using the images at full resolution (0.3m), we downsample the images with anti-aliasing by a factor of 2, to obtain images with a resolution of 0.6m. This is done because the sampling distance of the airborne LiDAR is 0.5m, meaning that we cannot evaluate the benefit of finer spatial resolution images.

We perform two sets of experiments on four different urban scenes of Jacksonville, shown in Figure~\ref{fig:albedo}. Each area covers 300$\times$300m.

The first experiment (Section \ref{sec:novel}) aims to evaluate how well S-NeRF can perform novel view synthesis in previously unseen conditions: both viewing and lighting.

The second experiment (Section \ref{sec:depth}) examines the accuracy of the surface shape by computing the expected altitude of the scene (Equation~\ref{eq:altitude}), evaluated with respect to the airborne LiDAR measurement.

\subsection{Novel view synthesis}
\label{sec:novel}
In Figure~\ref{fig:nerf} we compare novel views from unseen viewing and lighting conditions to real images in those conditions. Without the shadow model (NeRF), the shaded areas are smoothed together and form a grayish area without crisp edges. With our proposed S-NeRF model, the shadows are well positioned, and illuminated by a faint blue ambient sky light.

\begin{figure}
\begin{center}
   \includegraphics[width=.9\linewidth, page=5, trim={0 1cm 16cm 0.5cm}]{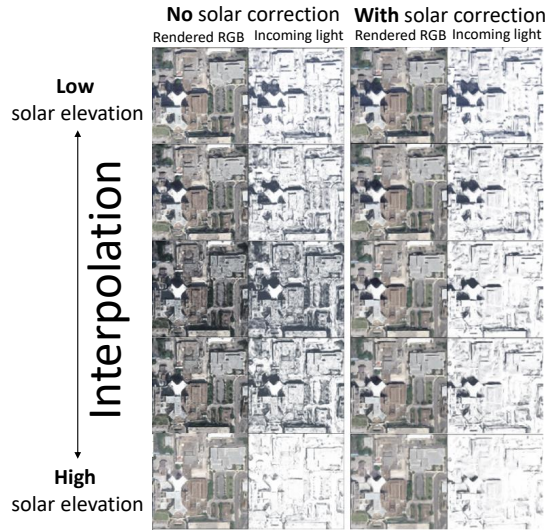}
\end{center}
\caption{\label{fig:interp}Interpolation of the solar direction vector $\mathbf{\omega_s}$ between two solar positions that are present in the training data. We compare two shading schemes. Left: with no solar correction, the incoming light is well estimated on training angles, but poorly estimated on unseen angles. Right: solar correction fixes darkening effects by learning the light source visibility based on the estimated scene geometry. These views are rendered at nadir to show how S-NeRF can produce synthetic ortho-images.}
\end{figure}

Figure~\ref{fig:interp} pushes this conclusion further by showing how S-NeRF allows for a smooth interpolation between two known lighting conditions. Because the RGB albedo values do not depend on the lighting conditions, the model maintains consistent colors when interpolating between different solar positions. The solar correction rays further allow for the shadows to be well placed and coherent with the estimated shape in previously unseen conditions. 

\begin{table}
\begin{center}
\begin{tabular}{lllll}
\hline
\multicolumn{1}{|l|}{Area index}      & \multicolumn{1}{l|}{004}            & \multicolumn{1}{l|}{068}            & \multicolumn{1}{l|}{214}            & \multicolumn{1}{l|}{260}            \\ \hline
\multicolumn{1}{|l|}{\# train/test}   & \multicolumn{1}{l|}{8/2}            & \multicolumn{1}{l|}{16/2}           & \multicolumn{1}{l|}{21/2}           & \multicolumn{1}{l|}{14/2}           \\ \hline
\multicolumn{1}{|l|}{Alt. bounds (m)} & \multicolumn{1}{l|}{0/-30}          & \multicolumn{1}{l|}{30/-30}         & \multicolumn{1}{l|}{80/-30}         & \multicolumn{1}{l|}{30/-30}         \\ \hline
\multicolumn{5}{c}{\textbf{SSIM (test set)}}                                                                                                                                                  \\ \hline
\multicolumn{1}{|l|}{NeRF}            & \multicolumn{1}{l|}{\textbf{0.364}} & \multicolumn{1}{l|}{\textbf{0.471}} & \multicolumn{1}{l|}{0.377}          & \multicolumn{1}{l|}{0.409} \\ \hline
\multicolumn{1}{|l|}{S-NeRF no SC}    & \multicolumn{1}{l|}{0.352}          & \multicolumn{1}{l|}{0.322}          & \multicolumn{1}{l|}{0.360}          & \multicolumn{1}{l|}{0.401} \\ \hline
\multicolumn{1}{|l|}{S-NeRF + SC}     & \multicolumn{1}{l|}{0.344}          & \multicolumn{1}{l|}{0.459}          & \multicolumn{1}{l|}{\textbf{0.384}} & \multicolumn{1}{l|}{\textbf{0.416}} \\ \hline
\multicolumn{5}{c}{\textbf{Altitude MAE (m)}}                                                                                                                                                 \\ \hline
\multicolumn{1}{|l|}{NeRF}            & \multicolumn{1}{l|}{5.607}          & \multicolumn{1}{l|}{7.627}          & \multicolumn{1}{l|}{8.035}          & \multicolumn{1}{l|}{11.97}          \\ \hline
\multicolumn{1}{|l|}{S-NeRF no SC}    & \multicolumn{1}{l|}{\textbf{3.342}} & \multicolumn{1}{l|}{4.799}          & \multicolumn{1}{l|}{\textbf{4.499}} & \multicolumn{1}{l|}{10.18}         \\ \hline
\multicolumn{1}{|l|}{S-NeRF + SC}     & \multicolumn{1}{l|}{4.418}          & \multicolumn{1}{l|}{\textbf{3.644}} & \multicolumn{1}{l|}{4.829}          & \multicolumn{1}{l|}{\textbf{7.173}} \\ \hline
\end{tabular}
\end{center}
\caption{\label{tab:results} Evaluation metrics on NeRF and S-NeRF. The upper table contains the number of training and test images used for each area, as well as the altitude bounds of the scene. Area indices correspond to the images in Figure~\ref{fig:albedo}. The middle table demonstrates the results of the novel view synthesis experiment. The best SSIM values are shown in bold. Overall, S-NeRF provides similar image quality to NeRF. The third table shows that S-NeRF always provides a lower altitude Mean Average Error (MAE) than NeRF. However, solar correction is not clearly beneficial for both metrics, meaning that the shading model and correction scheme could be improved further.}
\end{table}

\begin{figure}
\begin{center}
   \includegraphics[width=.9\linewidth, page=6, trim={0 3cm 18cm 0.5cm}]{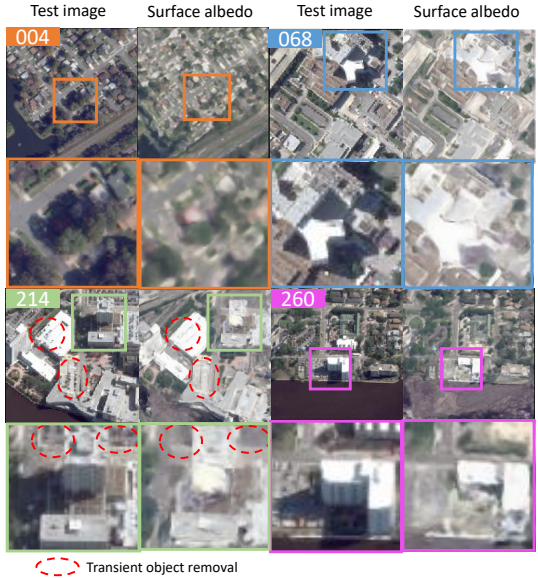}
\end{center}
\caption{\label{fig:albedo}Demonstration of S-NeRF shadow removal and transient object removal. The first and third column show unseen test images and the second and last column show the estimated albedo at the surface. The approach is mostly successful in restoring color to the persistent shadows. Remaining errors are particularly found at the intersection between the ground and vertical facades and along their outer edges.  We can also note the transient object removal in the three parking lots circled in area 214 (lower left).}
\end{figure}

In terms of quantitative results, Table~\ref{tab:results} shows that NeRF and S-NeRF produce statistically equivalent image quality scores, as measured by the SSIM. It appears as though in certain scenes, an overall darkening effect can counteract the increase in quality in the shadowed areas, which is visible in Figure~\ref{fig:nerf}. This table also demonstrates that solar correction maintains or increases the SSIM values, compared to when this scheme is not used, on 3/4 of the evaluated scenes. Area 004 (upper left image of Figure~\ref{fig:albedo}) is the only region where we observe a decrease in SSIM when using solar correction. We believe this is due to the small variations in altitude in this scene, which means that cast shadows occupy smaller parts of the images. Further statistical analysis is required to draw conclusions on the best way to apply solar correction.

Next, we show the estimated albedo at the surface for various scenes in Figure~\ref{fig:albedo}. This result shows how S-NeRF can be used to provide a 3D estimate of the albedo, even in areas that are entirely in the shadow in the training image set. This is because the model is able to learn the color of the ambient sky light, and its local contribution to the total perceived light. The performance cannot easily be quantified on real-world data as we have no widespread measure of the surface albedo.

We also observe that the albedo contains no or very few transient objects (specifically cars) as the model tries to learn a coherent representation for all of the training images. This is visible in area 214, on the lower left of Figure~\ref{fig:albedo}. The cars in the parking lots are mostly filtered out entirely, or at least blurred out. These outputs could be useful to detect transient objects in a new unseen image, or to provide a more robust semantic segmentation of the scene in the case where such objects are not of interest.

\subsection{Altitude extraction}
\label{sec:depth}
S-NeRF provides a full 3D model of the density from which~\cite{mildenhall2020nerf,sitzmann2020implicit} extract a surface boundary by selecting an iso-$\sigma$ surface and applying marching cubes. However, we lack a ground truth shape of the full 3D structure, so for this experiment we only evaluate the quality of the shape on its \emph{maximal altitude}. For this we render a vertical view and compute the Mean Average Error (MAE) between the expected altitude and the altitude measured by an airborne LiDAR~\cite{le2019ieee}.

\begin{equation}
\begin{split}
    \label{eq:altitude}
    \hat{h} & = \sum_{i=1}^{N_s} w_i h_i \\
    h_i & \in [h_{max}, h_{min}]\\
\end{split}
\end{equation}

Equation \ref{eq:altitude}~demonstrates how we estimate the surface altitude $\hat{h}$ from the volumetric representation. For each ray at a desired ground spatial resolution (0.5m), we sample altitudes $h_i$ between $h_{max}$ and $h_{min}$ (ray traveling downwards) using the sampling strategy described in Section~\ref{sec:altitude}. The estimated altitude is written as a weighted sum of the sample altitudes $h_i$ with weights $w_i$. By definition (Equation \ref{eq:alpha_comp}), $w$ is close to 1 at the location where the first object surface is visible.

Figure~\ref{fig:nerf} shows how S-NeRF corrects most of the shape inaccuracies that the original NeRF causes in shadowed areas. This is confirmed by Table~\ref{tab:results}, which indicates that S-NeRF produces a lower altitude MAE than NeRF on four different areas. In terms of comparison to Stereo Matching, we report that the average error of the predicted DEMs is in the same order of magnitude as the ones measured by~\cite{facciolo2017automatic}, which use a fusion of Stereo Matching predictions on 47 WorldView-3 images.

The training takes around 8 hours on a NVIDIA GeForce RTX GPU with 12GB RAM. We use an Adam optimizer with learning rate decaying from $1e^{-4}$ to $1e^{-5}$ over 100k iterations. The batch size is set to 256 pixel rays and 256 solar correction rays (when used). The number of samples and importance samples are 64 and 64, respectively.

\section{Discussion}
\label{sec:discussion}
One of the limitations of NeRF and S-NeRF alike is the need to retrain a neural network for each new scene. Recent efforts have been made towards generalizing neural volumetric representations to multiple scenes~\cite{schwarz2020graf, trevithick2020grf, yu2020pixelnerf}, and in making them more compact~\cite{liu2020neural}. This may be relevant for multi-view satellite imagery in the future. 

More robust transient object filtering and deeper temporal analysis could be attempted using latent embedding optimization~\cite{bojanowski2017optimizing}, taking inspiration from NeRF in the Wild~\cite{martin2020nerf}. This method can handle severe changes in illumination, but does not provide a solution to general relighting.

Neural Reflectance and Visibility Fields (NeRV)~\cite{srinivasan2020nerv} were very recently developed to address the issue of varying non-collocated light source conditions. This research was made in parallel to ours, so we did not perform a comparison with their method. They account for all sources of indirect illumination (including the scene itself) and model bidirectional specular effects, but require prior knowledge of the light conditions (sky color) and add a linear term to the complexity, which becomes $O(N+d)$ instead of $O(N)$.

\section{Conclusion}
This study proposes a novel adaptation of Neural Radiance Fields able to harness non-correlation effects (such as that of shadows) in satellite images and use them as a source of information rather than a source of perturbation with regards to 3D shape estimation.

To this end, an explicit light transport model is used to simultaneously take into account occlusion and varying illumination effects (shading). We propose to implicitly model the spectral irradiance with a learned light source visibility and ambient sky color, to resolve the ambiguities that this inverse problem implies while maintaining linear complexity.

Our experiments show that S-NeRF succeeds in generating realistic images in previously unseen viewing and lighting conditions, and in estimating the surface altitude with an accuracy of a few meters. Moreover, our methodology shows promise in detecting shadows, in recovering the albedo of surfaces with persistent shadows, and in localizing and removing transient objects.

\newpage
{\small
\bibliographystyle{ieee_fullname}
\bibliography{mybib}
}
\end{document}